\def\year{2020}\relax
\documentclass[letterpaper]{article} % DO NOT CHANGE THIS
\usepackage{aaai20}  % DO NOT CHANGE THIS
\usepackage{times}  % DO NOT CHANGE THIS
\usepackage{helvet} % DO NOT CHANGE THIS
\usepackage{courier}  % DO NOT CHANGE THIS
\usepackage[hyphens]{url}  % DO NOT CHANGE THIS
\usepackage{graphicx} % DO NOT CHANGE THIS
\usepackage{graphicx}  % DO NOT CHANGE THIS
\usepackage[utf8]{inputenc}
\usepackage[T1]{fontenc}

\usepackage{color}

\title{Automated Anonymisation of Visual and Audio Data in Classroom Studies}
\author{
\textbf{Ömer Sümer\textsuperscript{\rm 1}, Peter Gerjets\textsuperscript{\rm 2}, Ulrich Trautwein\textsuperscript{\rm 1}, Enkelejda Kasneci\textsuperscript{\rm 1}}\\
\textsuperscript{\rm 1} University of Tübingen\\  
\textsuperscript{\rm 2} Leibniz-Institut für Wissensmedien \\ 
Tübingen, Germany\\
\{oemer.suemer,ulrich.trautwein,enkelejda.kasneci\}@uni-tuebingen.de\\
p.gerjets@iwm-tuebingen.de
}

\begin{document}

\maketitle

\begin{abstract}
Understanding students' and teachers' verbal and non-verbal behaviours during instruction may help infer valuable information regarding the quality of teaching. In education research, there have been many studies that aim to measure students' attentional focus on learning-related tasks: Based on audio-visual recordings and manual or automated ratings of behaviours of teachers and students. Student data is, however, highly sensitive. Therefore, ensuring high standards of data protection and privacy has the utmost importance in current practices. For example, in the context of teaching management studies, data collection is carried out with the consent of pupils, parents, teachers and school administrations. Nevertheless, there may often be students whose data cannot be used for research purposes. Excluding these students from the classroom is an unnatural intrusion into the organisation of the classroom. A possible solution would be to request permission to record the audio-visual recordings of all students (including those who do not voluntarily participate in the study) and to anonymise their data. Yet, the manual anonymisation of audio-visual data is very demanding. In this study, we examine the use of artificial intelligence methods to automatically anonymise the visual and audio data of a particular person. 
\end{abstract}

\section{Introduction}\label{section:introduction}
\noindent Visual and audio recording of classroom instructions has been widely used in education research for many purposes such as self-reflection, peer collaboration, teacher coaching, or classroom observation research for assessment and evaluation of teaching quality. Besides the traditional approaches in classroom observation and behaviour coding systems \cite{Helmke:1992}, there are recent efforts in multimodal learning analytics that aim to automate these workflows \cite{Goldberg:2019,Suemer:2018}. The progress in the area of machine learning and artificial intelligence have additionally paved the way to conduct classroom studies in large-scale and automate these analyses. Unlike traditional types of data such as questionnaires or written reports, audio-visual recordings of a classroom cannot be easily anonymised as they lose the information which is required for further manual or automated analysis.

A typical practical challenge during such studies is that there may be students who do not want to consent in the study or their data being used by education practitioners. A common approach to overcome this limitation is to reorder the classroom by changing the seats of these persons so they can sit out of the video coverage. However, this constitutes an unnatural intervention in the usual form of classroom organisation. Furthermore, the students who do not participate in the study may raise their hands and actively participate in the ongoing instruction by speaking. Since voice also contains private information, the collection, storage, distribution, and analysis of this kind of data might cause severe violations of data protection laws and regulations.

In fact, the current digital transformation of our society goes along with a corresponding change and regulation of data protection laws. Whereas the US data protection legislation is developing and, in comparison, Germany has the most comprehensive and the first data protection law in the world (Hessen, 1970) \cite{Hoel:2018}. In Europe, the General Data Protection Regulation (GDPR) of the European Union came into effect in May 2018. %Not only restricted to EU, but it also led many companies from other countries that collect, store and process personal data of European citizens to revisit their data protection practice. 
According to DLA Piper\footnote{\url{https://www.dlapiperdataprotection.com/}}, there are also several Asian countries which recently passed data protection laws. There is also an increasing interest in the privacy of personal data, independent of the underlying societal and individual foundations and the scope of regulations.

In August 2019, Sweden issued its first fine to a public board as they used facial recognition technology to keep track of class participation during a few weeks in a pilot study that aimed to automate the class register.  Even though the data was collected and stored in local and locked computers without internet connection,  according to the Swedish data protection agency, they violated the GDPR in three ways: (i) by processing personal data in a more intrusive manner than what was necessary for the purpose (monitoring of attendance), (ii) processing sensitive personal data without legal basis, and (iii) not fulfilling the requirements of data protection impact assessment and prior consultation. \cite{Swedish_DPA:2019}.
The Swedish example shows that even a simple use of visionary technologies may cause data breaches and that data protection in educational situations is highly critical and must be carried out after prior consultation. Contrary to the common understanding that computer vision and machine learning are used to collect and analyse private, personal data, we show that by leveraging these technologies, multimodal data can be efficiently and semi-automatically anonymised.

Let us consider an audio-visual recording of teaching situations at schools, which is typical research means in the field of classroom management since the 1990s. Under the national data protection laws and the GDPR, the current practice is to pseudo-anonymise the questionnaire data (pre- and post-tests). Pseudo-anonymisation of categorical data can be done using code words assigned to students, instead of using their names. However, audio-visual data cannot be easily anonymised because they need to either be watched by education researchers or processed by computer algorithms.

In studies that necessitates audio-visual data collection in the classroom or small group discussions, there may be subjects who do not want to participate in the research. Excluding these students may be impractical and raise other complicated issues. In contrast to longer storage and data processing, we propose to collect and store classroom data for only a short period (a few hours to max. one day). During this time, the collected audio-visual data can be automatically processed, anonymised, and validated immediately. 

\textbf{Our contributions.} We create a small-scale classroom observation dataset that is composed of 6,5 hours of instruction in varying subjects and show the feasibility of automated anonymisation approaches on audio-visual data. Even though our focus is educational data such as classroom observation recordings our approach is applicable to any other domain of audio-visual data and makes the anonymisation much more effective.

\section{Related Works}\label{section:relatedworks}
In this section, we will review the previous works that concern data anonymisation. As we mainly aim at audio-visual data anonymisation, our focus will be video and audio.

\paragraph*{Video. } Faces are perhaps the most characteristic features containing private information in video data. In the domain of computer vision, face detection, tracking, and recognition are among the widely studied problems. In the recent years, open and large-scale databases and the progress in the field of deep learning have led to fast and accurate results for face detection, tracking and recognition even under challenging conditions such as occlusion, lightning chances, eyeglasses, or make-up. 

In categorical datasets, the injection of noise using differential privacy techniques is the right choice to protect private information \cite{Ji:2014}. However, these methods are not suitable for images, and in particular, for face images since the injected noise deteriorates the required face features. Even though it is far from guaranteeing privacy protection, crafting adversarial samples \cite{Goodfellow:2015} on images can at least ensure many trained face recognisers fail on perturbed images. However, the anonymisation of faces is not only addressing privacy protection for face identification algorithms but also face recognition by humans.

 In recent years, there have been many successful applications of generative models to anonymise faces, such as adversarial generative networks and variational autoencoders. For example, a face can be swapped with another person's face to hide the original identity. In \cite{Ren:2018}, a face modifier network was trained together with adversarial regulariser and an action recognition network, whose task is to recognise the action even in the modified face. In this way, anonymised data can be used to train action recognition models. In contrast, \cite{Hukkelas:2019} first filled the face that has to be anonymised with noise and applied then a conditional GAN using these patches and facial keypoints. In this manner, the authors constrained the background and head pose and guaranteed not to have any leakage of private information.

The audio-visual data is used for either automated analysis or manual inspection of behavioural data. We found two drawbacks of anonymisation based on generative models. First, they do not ensure temporally coherent generations and may induce flickering identities in some participants, which distracts the viewer who uses the material for educational purposes. Second, a face that looks like another identity does not guarantee that details in facial expressions are preserved.

In this paper, we are addressing the face in the visual domain. However, beyond faces, there are other features such as hair clothing, and body pose and they may contain private information. In the discussion section, we present how to fulfil different levels of anonymisation.

\paragraph*{Audio.} The task of audio anonymisation aims to identify all intervals of a participant whose voice is audible. This task can be approached using speaker diarisation. In general, diarisation is composed of several sub-tasks such as speaker activity detection, speaker change detection, extraction of an embedding representation and clustering. After having speaker diarisation, we acquire clusters of speaking patterns where each of them belongs to a different participant. Then, a possible way is to ask a human annotator to locate an exemplary of the anonymised person's voice. Eventually, we can anonymise all parts assigned to the same cluster. 

The focus in this field is to develop better embeddings, which in practice can be acquired from the internal representations of speaker recognition networks as in d-vectors \cite{Variani:2014} and \cite{Garcia-Romero:2017}. Recently, \cite{Zhang:2019} modelled speakers using multiple instances of a parameter-sharing recurrent neural network. In contrast to previous works based on clustering, they identified different speakers interleaved in time. 

\begin{figure*}[ht!]
\centering
\includegraphics[width=0.95\textwidth]{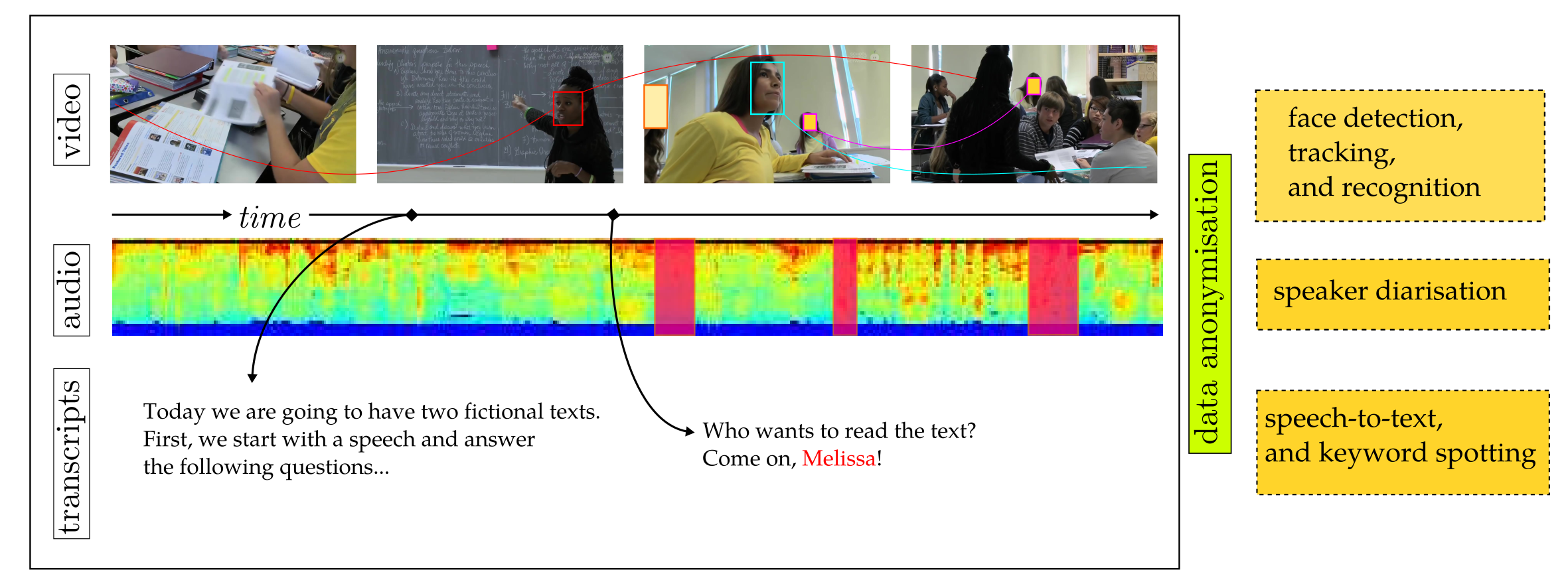} 
\caption{Our proposed anonymisation pipeline on multimodal educational data. }
\label{figure:01}
\end{figure*}

Similar to changing faces with a fake identity, there exist various anonymisation approaches in the context of audio analysis. For example, \cite{Fuming:2019} extracted speaker identity features from an utterance and synthesised the input by replacing the identity part with a pseudo-identity. 

The most important part is to locate the speaking patterns of a person. Later, there are two possible alternatives: either to synthesise the voice by removing private acoustic parameters or directly silence these parts. We prefer the latter one because when they did not give consent to participate in the study, storage of behavioural data would not be an option even by removing identity information. Furthermore, there can be private cues in the content of their speech.

\section{Approach}\label{section:approach}
In practice, we have two working scenarios. (i) We are given one or several participants to anonymise their data, and (ii) we are being asked to anonymise all student data except for the teacher's instruction (e.g. when the teacher's instruction is used for rhetorical analysis of teacher). Our audio-visual anonymisation workflow is based on multiple machine learning steps and shown in Figure~\ref{figure:01}. The modalities that contain private information in our application are video and audio (including the spoken information during the conversations in the learning context). Depending on the level of anonymisation expected, there can be varying levels of anonymisation in visual and audio data. In this section, we describe our approach to anonymise faces in the videos and speaking patterns in the audio recordings.

\paragraph*{Anonymising faces.} Videos can be taken from handheld or static cameras. Also, they can contain footage taken from different cameras. The first step is, therefore, to locate shot boundaries and create scenes. To detect shot boundaries, including hard cuts and gradual transitions, we used the TransNet \cite{Soucek:2019} approach, which is based on 3D dilated CNNs.

After detecting shot boundaries, we can define scenes, $s_i=[I^1,...,I^{N_i}]$, where each scene $s_i$ can be in varying length, either a few seconds or minutes. In each scene, we employ the RetinaFace face detector \cite{Deng:2019} which is trained on the WIDER face dataset \cite{Yang:2016}. On the contrary to the face detection methods in the literature that require several stages of networks, RetineFace is trained using multiple losses which combines extra-supervision and self-supervised multitask learning objectives. These additional losses include face classification, bounding box regression, extra supervision of facial landmark regression, and self-supervised mesh decoder for predicting a pixel-wise 3D shape face.

We acquire a set of faces with bounding boxes $\{[b^1_1,...,b^1_{n_1}],...,[b^t_1,...,b^t_{n_t}] \}$, where $t$ is the index of frames that belong to the scene, $s_i$, and $n_t$ is a varying number of faces detected in each frame. Either in small groups or traditional classroom formation often students sit in their usual seats and do not change their seats in a scene. Furthermore, the face detector shows very decent performance even under challenging situations, such as motion blur, partial occlusion, or low image resolution. Thus, we employ a location-based tracking approach. 

We first calculate the Intersection over Union (IoU) scores between the bounding boxes in the consecutive frames:
\begin{equation}
    s(b_t, b_{t+1})= \frac{b_t \cap b_{t+1}}{b_t \cup b_{t+1}}
\end{equation}
where $b=[x,y,w,h]$'s are bounding boxes. Then, we convert the IoU scores to a cost matrix and assign the faces in the current frame to the next one using minimum weight matching in bipartite graphs that is also known as Hungarian algorithm \cite{Kuhn:1955}. In this way, we find correspondences that minimize the cost between the detections of consecutive frames. As we maximize the intersection between bounding boxes of the same faces, the IoU scores are negated. Even though a combination of several cost terms (i.e. additionally visual similarity between faces) can yield more reliable tracklets, in practice, we observed that the location cue is enough to create tracklets in classroom scenes.

\begin{figure*}[ht!]
\centering
\includegraphics[width=0.95\textwidth]{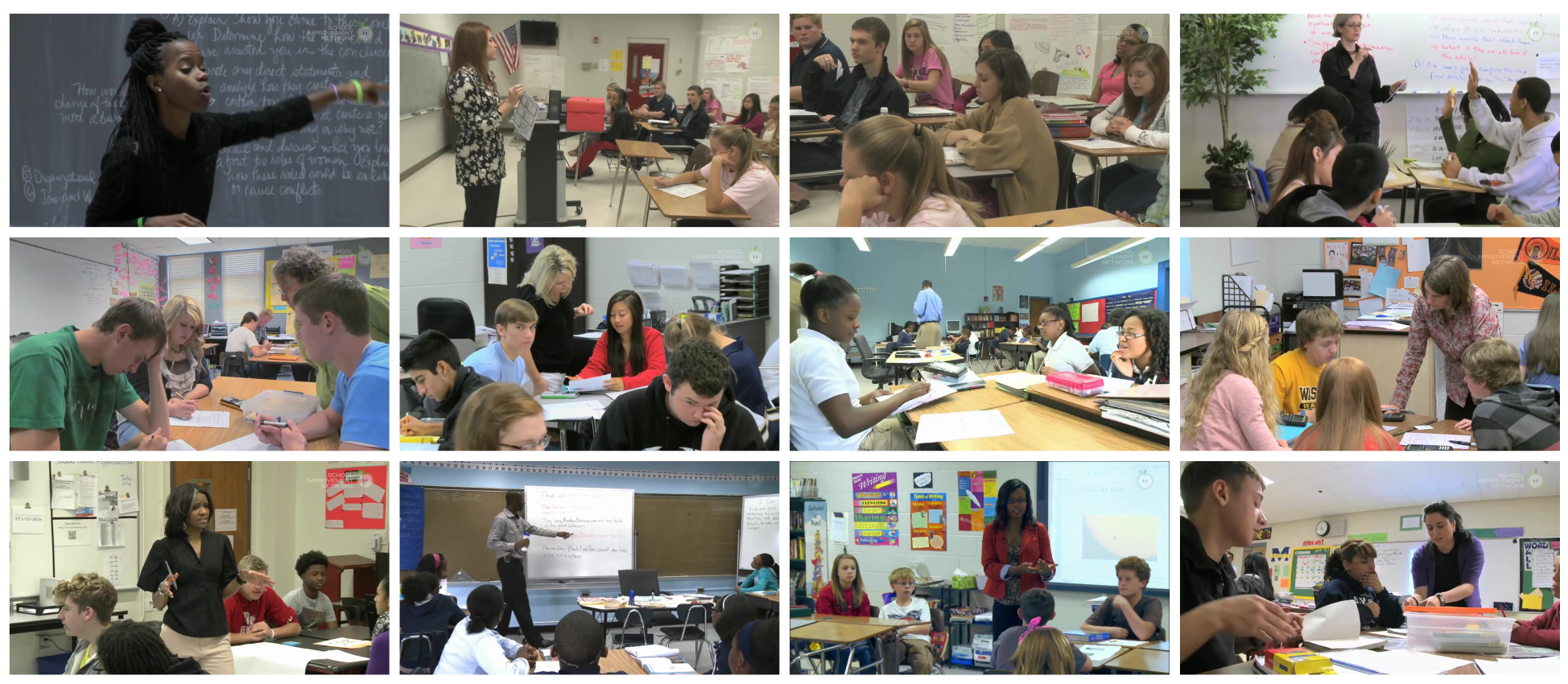} 
\caption{Snapshots from the classroom observation videos.}
\label{figure:samples}
\end{figure*}

After having face tracks, we pick only one sample of the person that we want to anonymise and use this sample as a reference for the further face verification steps. Depending on the camera angle and the subjects' movement, their head poses may show significant variation. Thus, it is crucial to pick a face track that represents these variations, for instance, the one longer and containing different head poses. As a face embedding, we use the Inception ResNet (V1) trained minimizing Triplet loss \cite{Schroff:2015} on the VGGFace2 dataset \cite{Cao:2018}. Embedding vectors are $L_2$ normalised and their dimension is 512. To acquire the similarity of any face tracks in the video with respect to our reference, we use cosine similarities between them:

\begin{equation}
score = \min \frac{F(I_{(b_i,t_i)}) \cdot F(I_{(b_j,t_j)}) }{ ||F(I_{(b_i,t_i)})|| \  ||F(I_{(b_j,t_j)})|| }
\end{equation} 
where $F(.)$ is the face embedding, $I_{b_i,t_i} \in \mathcal{T}_{(test)}$ and $I_{b_j,t_j}$ is any face track from the same video where we want to find a specific subject whose face has to be anonymised. Eventually, we threshold the cosine similarity scores to retrieve query identity and either make these faces blurry or set them to black.\\

\textbf{Finding speaking patterns. } Similar to the face information, human voice also contains biometric traits. The main task in our context is to find the speaking patterns of a participant whose data has to be anonymised. We use the Unbounded Interleaved-State Recurrent Neural Network (UIS-RNN) \cite{Zhang:2019} to diarise speaking patterns of different speakers.

An unbounded RNN network, the UIS-RNN,  fed with d-vectors that is extracted from equal-length audio segments. The number of speakers is decided using a Bayesian non-parametric process, and different speakers are modelled within the states of RNN in the time domain. As the UIS-RNN approach performed better than k-means and Spectral Clustering of d-vectors on benchmark datasets, we address speaker diarisation using a pre-trained UIS-RNN model. 

When we apply diarisazion to classroom data, we deploy the speaker diarisation to the spectrograms of the entire classroom recordings. Subsequently, we show these clustered audio parts to annotators together with the original video and ask them to pick the cluster to be anonymised.

\begin{figure}[b!]
\centering
\includegraphics[width=0.95\columnwidth]{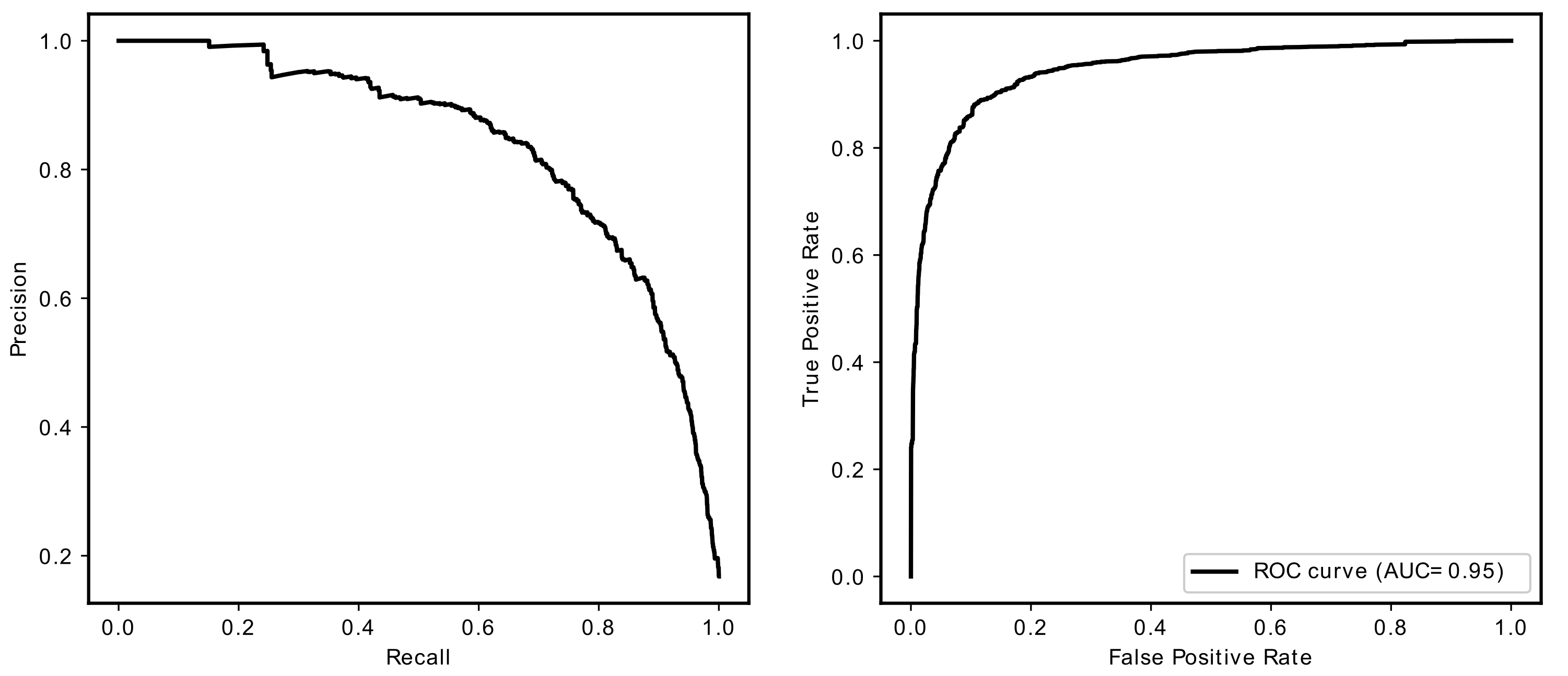} 
\caption{Face verification results on classroom observation videos: Precision vs. Recall and ROC curves.}
\label{figure:results_face}
\end{figure}

\section{Experimental Results}\label{section:experimentalresults}
In order to test the feasibility of the automated anonymisation in classroom observation videos, we first created a collection of classroom observation videos from YouTube. These videos are either recorded from a static point or using handheld cameras and contain shot transitions. More specifically, the video material consists of 18 classroom instruction videos, where the taught topics include English language arts, maths, and social studies.  The total duration of this instruction material is approximately 6.5 hours. All videos are available in the resolution of 1280x720px. Sample keyframes are depicted in Figure~\ref{figure:samples}.

In the face anonymisation task, we first created all face tracks. Then, on 14 participants where their faces are visible in different parts of videos, we manually labelled whether they belong to the selected test identity or not. The evaluation task is to pick a representative face track and calculate the similarity of each face track with respect to the query. 

Figure~\ref{figure:results_face} shows the test results in terms of precision vs recall and receiver operating curves (ROC) per image. The Area under Curve (AUC) is 0.95, which means that our approach can recognise a given query identity with a high precision.  In the performance of face identification of anonymised participant, we observed that the effect of face tracking is essential because it makes the verification more stable. Besides, using a representation that is trained on a dataset that contains large pose variations is also helpful.    

\begin{figure}[ht!]
\centering
\includegraphics[width=\columnwidth]{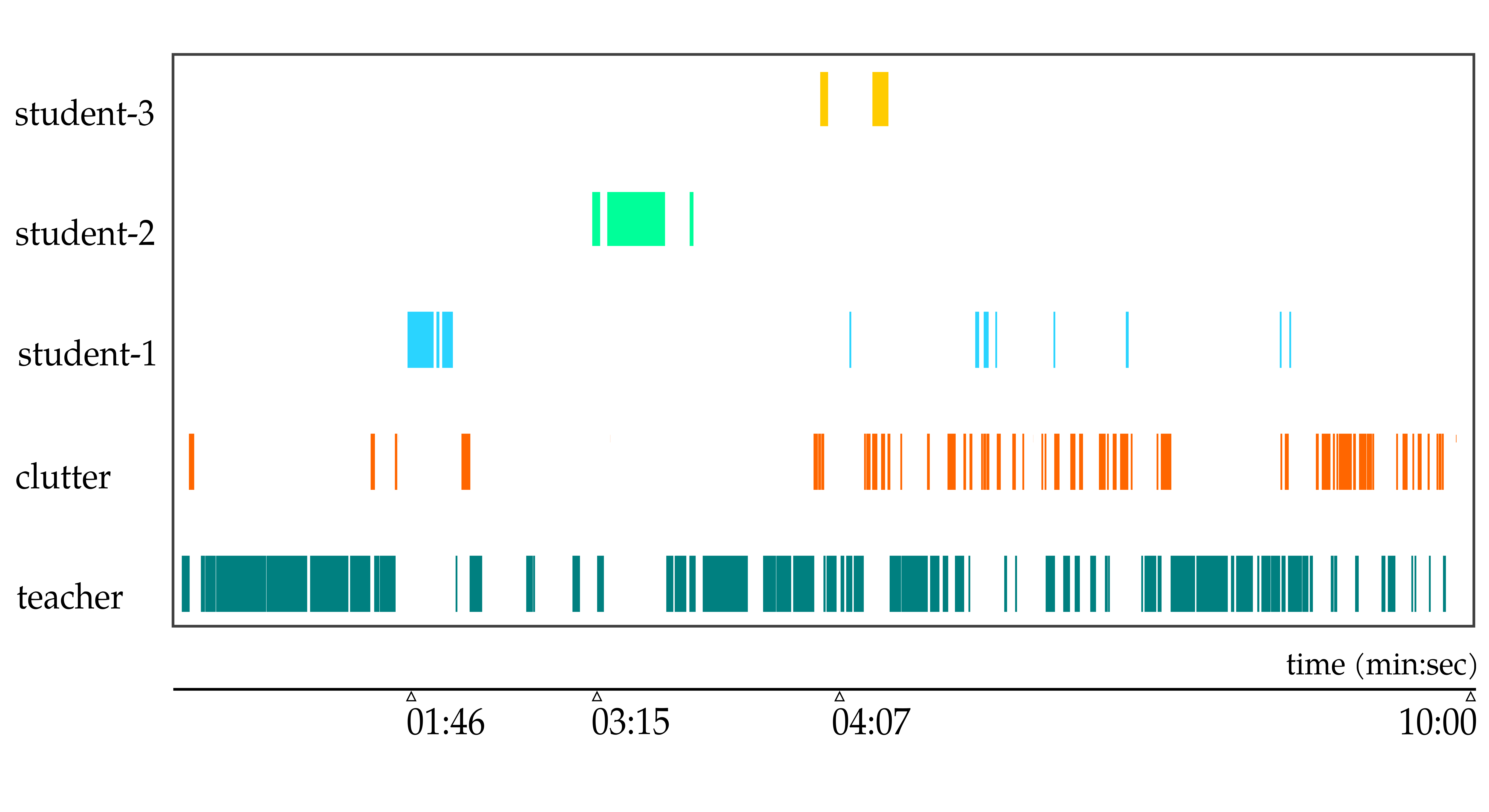} 
\caption{An example output of speaker diarisation on a 10-minute classroom instruction from the dataset.}
\label{figure:results_audio}
\end{figure}

With regard to the audio analysis, we segmented the audio signal of an example instruction scene from our dataset. The output of the speaker diarisation can be seen in Figure~\ref{figure:results_audio}. In this context, the UIS-RNN approach is quite robust to find the speaking patterns of a participant; however, in audio sequences longer than 15-20 minutes, we observed that it might over-segment some speakers, particularly the one speaking most of the time. 

In order to spot the speaking pattern of a specific speaker on a long recording, such as whole day instruction with several teaching units, there are two alternative approaches. One is to use speaker diarisation and manually picking the anonymised identity in shortened clips of the audio recordings. This option requires more human effort to prove and check the quality of the diarisation. Alternatively, we may ask an annotator to pick a few samples of the anonymised person from different parts of the recording and retrieve most likely parts with respect to d-vectors. We observed that the presence of a face in the current frame is also a further precursor of the active speaker can be that person.

In addition to anonymisation, the  speaking patterns of a classroom also indicate how a particular student actively participates in the learning processes during the class. 

\section{Discussion and Future Outlook}\label{section:discussion}
In this section, we first describe the levels of data anonymisation in classroom studies, summarise then our key findings and give an outlook on our future work in this context. 

\subsection{Different levels of data anonymisation}
We focus on the automated anonymisation of video (faces) and audio data in the educational context. In practice, the level of data anonymisation and needed modalities may vary. In this part, we first address the modalities in detail and then propose different levels of anonymisation.\\

\textbf{Questionnaires. }Many educational studies, including the ones that require video recording, use questionnaires. They may contain questions regarding personal information, educational, or socio-economical background of students. Particularly personal data, such as biometric data, rate, ethnicity, and political opinion requires the highest level of protection. The most common approach to this challenge is pseudonymisation. 

It should be noted, however, that the pseudonymised data may still be vulnerable to re-identification and remains therefore personal. In order to prevent re-identification, differential privacy is the current practice \cite{Francis:2019} and should be applied not only to data of specific participants who do not want to participate in the study, but to be considered for all participants.\\

\textbf{Video. }Faces in the videos are the primary modality that contains personal information. Therefore, our focus here was  on finding and anonymising faces. However, there are other soft biometric traits \cite{Jain:2004} which may require anonymisation. Some examples are hair colour, body posture, and gesture patterns. When the face of an anonymised person is detected and recognised, hair can be occluded, for example by extending the face bounding box around and top of the face.

So far, we did not consider body and gestures in our analysis since this can be done by employing human pose estimation. In case of a classroom recording including approximately 20 students, deploying a multi-person pose estimation method can considerably increase the processing time. However, we will exploit this issue in our future work.\\

\textbf{Audio. }In our analysis, we tested speaker diarisation. Notably, the audio domain requires a more thorough understanding of the spoken conversation. For instance, even if we anonymised face and speaking patterns of a participant, other participants may talk about the anonymised person or content. In order to address this issue, we need a semantic understanding of the spoken text using speech-to-text approaches and efficient search techniques for the information relevant to selected keywords on the entire data. 

In a manner that each level contains the anonymisation of previous modalities, we can categorise the varying levels of anonymisation for classroom studies as follows:

\begin{enumerate}
    \item Pseudonymisation of questionnaires and personal data,
    \item Anonymisation of hard biometric traits such as faces in the video and speaking time,
    \item Anonymisation of  soft biometric traits such as body posture, gestures, and hair appearance,
    \item Anonymisation of the spoken information during the instruction unit (either by anonymised participants or other persons) such as students' and school's name or location.
\end{enumerate}

The appropriate level of anonymisation can be chosen according to the nature of the data and institutional board review approval regarding ethical aspects of data collection, storage and processing.

We can summarise the key findings of our work as follows:
\begin{itemize}
    \item We created a small classroom observation dataset that is representative of typical scenarios in educational studies.
    \item We investigated the feasibility of automated anonymisation on visual and audio domains. The essential modality on the visual domain is anonymisation of faces, for which we showed that it could be solved in the context of classroom observation data at high accuracy. Similarly, a speaker diarisation approach helps to find speaking clusters and those parts in the audio signal belonging to pointing to participants that have to be anonymised. 
\end{itemize}

As it is the case in most vision and machine learning applications, none of the methods can guarantee completely accurate predictions. Furthermore, the owner of the data will be ethically and legally responsible for the proper and accurate anonymisation of the visual and audio modalities. Our proposed approach can be seen as an alternative to speed up anonymisation in a way that requires only quick manual inspection and correction.  

For further manual inspection, we can export the output of face identification and speaker diarisation results in a compatible format with common multimedia annotation tools such as the VGG Image Annotator \cite{Dutta:2019} or the ELAN software \cite{Sloetjes:2008}.

In future work, we plan to create a simple interface that is compatible with available multimedia annotation tools and can be used by educational practitioners without any technical experience and to conduct a user study on real multimodal collection and analysis process of educational data.

\paragraph*{Acknowledgements} Ömer Sümer is a doctoral student at the LEAD Graduate School \& Research Network [GSC1028], funded by the Excellence Initiative of the German federal and state governments. This work is also supported by Leibniz-WissenschaftsCampus Tübingen ``Cognitive Interfaces''.

\bibliography{references}
\bibliographystyle{aaai}

\end{document}